\documentclass{article}

\usepackage{PRIMEarxiv}

\usepackage[utf8]{inputenc} 
\usepackage[T1]{fontenc}    
\usepackage{hyperref}       
\usepackage{url}            
\usepackage{booktabs}       
\usepackage{amsfonts}       
\usepackage{nicefrac}       
\usepackage{microtype}      
\usepackage{lipsum}
\usepackage{fancyhdr}       
\usepackage{graphicx}       
\graphicspath{{media/}}     
\usepackage{siunitx}
\usepackage{caption}
\usepackage{subcaption}
\usepackage{amsmath,amssymb,amsfonts}%

\title{Autonomous Navigation of Catheters and Guidewires in Mechanical Thrombectomy using Inverse Reinforcement Learning
\thanks{\textit{\underline{Citation}}: 
\textbf{Robertshaw, H., Karstensen, L., Jackson, B. et al. Autonomous navigation of catheters and guidewires in mechanical thrombectomy using inverse reinforcement learning. Int J CARS (2024). https://doi.org/10.1007/s11548-024-03208-w}} 
}

\author{
  Harry Robertshaw, Benjamin Jackson, Alejandro Granados, Thomas C Booth \\
  School of Biomedical Engineering \& Imaging Sciences \\
  Kings College London \\
  London\\
   \And
  Lennart Karstensen \\
  AIBE \\
  Friedrich-Alexander University Erlangen-Nürnberg \\
  Erlangen\\
}

\begin{document}
\maketitle

\begin{abstract}
    \textbf{Purpose:} Autonomous navigation of catheters and guidewires can enhance endovascular surgery safety and efficacy, reducing procedure times and operator radiation exposure. Integrating tele-operated robotics could widen access to time-sensitive emergency procedures like mechanical thrombectomy (MT). Reinforcement learning (RL) shows potential in endovascular navigation, yet its application encounters challenges without a reward signal. This study explores the viability of autonomous guidewire navigation in MT vasculature using inverse reinforcement learning (IRL) to leverage expert demonstrations.
    
    \textbf{Methods:} Employing the Simulation Open Framework Architecture (SOFA), this study established a simulation-based training and evaluation environment for MT navigation. We used IRL to infer reward functions from expert behavior when navigating a guidewire and catheter. We utilized the soft actor-critic algorithm to train models with various reward functions, and compared their performance \textit{in silico}.
    
    \textbf{Results:} We demonstrated feasibility of navigation using IRL. When evaluating single versus dual device (i.e., guidewire versus catheter and guidewire) tracking, both methods achieved high success rates of 95\% and 96\%, respectively. Dual-tracking, however, utilized both devices mimicking an expert. A success rate of 100\% and procedure time of 22.6\,\unit{\second} was obtained when training with a reward function obtained through `reward shaping'. This outperformed a dense reward function (96\%, 24.9\,\unit{\second}) and an IRL-derived reward function (48\%, 59.2\,\unit{\second}).
    
    \textbf{Conclusions:} We have contributed to the advancement of autonomous endovascular intervention navigation, particularly MT, by effectively employing IRL based on demonstrator expertise. The results underscore the potential of using reward shaping to efficiently train models, offering a promising avenue for enhancing the accessibility and precision of MT procedures. We envisage that future research can extend our methodology to diverse anatomical structures to enhance generalizability.
    
\end{abstract}

\keywords{Inverse reinforcement learning \and Mechanical thrombectomy \and Machine learning \and Artificial intelligence \and Autonomous navigation \and Endovascular intervention}

\section{Introduction}\label{sec1}

    Cardiovascular (CV) diseases are the most common cause of death across Europe, accounting for more than 4~million deaths each year, with cerebrovascular disease accounting for 25.4\% of CV-related mortality across all ages and genders \cite{Townsend2016}. Mechanical thrombectomy (MT) has become an established treatment for patients with acute ischemic stroke due to large vessel occlusion, increasing the likelihood that a patient will be functionally independent after a stroke \cite{Goyal2016, Vidale2017, Rha2007, Berkhemer2015}. During such a procedure, an operator navigates a guidewire and guide catheter from an insertion point (typically the common femoral artery) through the common iliac artery and descending aorta to the aortic arch. Subsequently, a guide catheter is typically advanced over a guidewire through the aortic arch to the distal aspect of the cervical segment of the internal carotid artery (ICA) (with or without the help of a slip catheter). Once the guide catheter is positioned in the distal aspect of the cervical segment of the ICA it serves as an anchor for a subsequent step. The subsequent step is often the navigation of a micro-guidewire and microcatheter through the distal aspect of the cervical segment of the ICA to the site of the thrombus which is typically the M1 segment of the middle cerebral artery (MCA). Once the target site has been reached, a mesh-like device called a stent retriever is pushed through the thrombus within a microcatheter, expanded to engage the thrombus, and then pulled backwards, which removes the thrombus from the blood vessel. Alternatively, an aspiration catheter – where a vacuum sucks the thrombus from the artery - is used instead of a stent retriever.
    
    In acute ischemic stroke, time from symptom onset to treatment is crucial, as the benefits of MT become more marked the sooner a thrombus is removed \cite{Saver2016}. As a result, in the UK for example, only 3.1\% of stroke admissions benefit from MT despite at least 10\% of patients eligible for treatment \cite{McMeekin2017, SSNAP2023}. Other challenges for MT relate to occasional complications, including perforation, thrombosis and dissection in the parent artery, and distal embolization of thrombus \cite{Hausegger2001}. Moreover, angiography requires intravascular contrast agent administration, which can occasionally lead to nephrotoxicity \cite{Rudnick1995}. For operators and their teams, the high cumulative dose of x-ray radiation from angiography is a risk factor for cancer and cataracts \cite{Klein2009}. Although exposure can be minimized with current radiation protection practice, some measures involve operators wearing heavy protective equipment, a risk factor for orthopedic complications, so alternative exposure reduction methods are beneficial \cite{Ho2007, Madder2017}.

    It is hoped that robotic surgical systems can mitigate or eliminate some of the challenges that MT presents. For example, robotic systems could be set up in hospitals nationwide and tele-operated remotely from a central location, increasing the speed of access to treatments such as MT beyond what is currently possible \cite{Crinnion2022}. Additionally, robotic systems might eliminate any operator physiological tremors or fatigue and allow MT to be performed in an optimum ergonomic position while potentially increasing procedural precision (for example, procedure time), thereby improving overall performance and reducing complication rates \cite{Riga2010}. Furthermore, as operators would not be required to stand next to the patient, their radiation exposure would be reduced, and the need to wear heavy protective equipment would be eliminated.

    Robotic systems such as the \textit{Magellan}\textsuperscript{TM} system (Auris Health, Redwood City, USA) and the \textit{Corpath GRX}\textsuperscript{\textregistered} (Corindus Vascular Robotics, USA) have been used to help alleviate some of the challenges of MT, however, they have limitations. The controller-operator structure requires a reasonably high cognitive workload and can still result in human error which means that the procedure is limited to an individual operator's skill set in terms of both MT and robot handling \cite{Mofatteh2021}. These robotic systems consist of user interfaces such as buttons and joysticks, requiring skills different from those used in current clinical practice \cite{Jackson2023}. Additionally, a lack of haptic feedback from robotic systems results in difficulty receiving tactile feedback from the catheters and guidewires as they interact with vessel walls \cite{Crinnion2022}.

    One emerging method of mitigating some of these challenges is by applying artificial intelligence (AI) techniques to robotic systems. AI, and in particular, machine learning (ML), has accelerated in recent years in its applications for data analysis and learning \cite{Sarker2021}, with many areas of healthcare already making use of this technology for disease prediction and diagnosis \cite{Fatima2017, Silahtaroglu2021}. By using ML in the autonomous navigation of guidewires and catheters in MT, it is plausible that in endovascular specialities facing a shortage of highly-trained operators, tele-operated MT may be performed safely and effectively by a few highly-trained operators based in centralized neuroscience centres. Alternatively, less experienced operators – for example, endovascular operators (e.g., interventional radiologists who are not used to performing neurointervetional procedures) may use AI-assisted robots in hospitals that are not neuroscience centres (the majority of hospitals are not neuroscience centres). Potentially, such developments would lead to greater accessibility of MT globally \cite{Robertshaw2023}.

    Several papers have investigated the potential use of ML in the automation of catheters and guidewires for endovascular interventions, with a recent systematic review finding 80\% (8/10) of studies (all published after 2018) implemented some form of reinforcement learning (RL) \cite{Robertshaw2023}. RL is a subset of ML where an agent learns by interacting with the environment, receiving feedback as rewards, and aims to minimize cumulative rewards over time by optimising its actions based on the current state \cite{Sutton1998}. However, this trial-and-error approach means that RL often requires many interactions with the environment to learn the optimal policy \cite{Adams2022}. `Demonstrator data' has been utilized previously in a variety of ways: it has been used to establish control policies for RL to optimize \cite{Chi2018}, enhance training speed \cite{Behr2019}, and also function as high-priority samples in the reinforcement learning replay memory \cite{Kweon2021}. While a `dense reward', characterized by frequent feedback, has been shown to provide quicker training times than a `sparse reward', which offers feedback less frequently \cite{Behr2019}, there has been no previous work to use demonstrator data to determine a reward function specifically for autonomous endovascular navigation tasks. Inverse reinforcement learning (IRL) is able to do this by using an agent to infer the underlying reward function from the observed behaviour of an expert \cite{Ng2000}. By leveraging the knowledge of experts, the number of trial and error cycles needed can potentially be reduced, leading to faster and more effective learning of complex tasks.
    
    This study aimed to leverage expert demonstrators through IRL to train autonomous guidewire and guide catheter navigation in simulated MT environments (i.e., \textit{in silico}), by navigating from the common iliac artery to the distal aspect of the cervical segment of the ICA (representing the early stages of navigation which is completed by obtaining this stable guide catheter position). A primary objective was to train and test three models, each with a different reward function, using the soft actor-critic (SAC) algorithm \cite{Haarnoja2018}. A secondary objective was to conduct an analysis of one and two-device tracking. This study contributes to the advancement of autonomous MT navigation and demonstrates a novel application of IRL in the simulated MT vasculature environment, filling a significant research gap in the field \cite{Robertshaw2023}.

\section{Methods}\label{sec2}

    In the following sections, we present the navigation task, the environment required to simulate MT and the simulation data collection process of expert demonstrator data. Subsequently, we describe the IRL approach for devising a reward function.

    \subsection{Navigation Task}\label{navigation}

\begin{figure}
            \centering
            \begin{subfigure}[b]{0.3\textwidth}
                 \includegraphics[width=\textwidth]{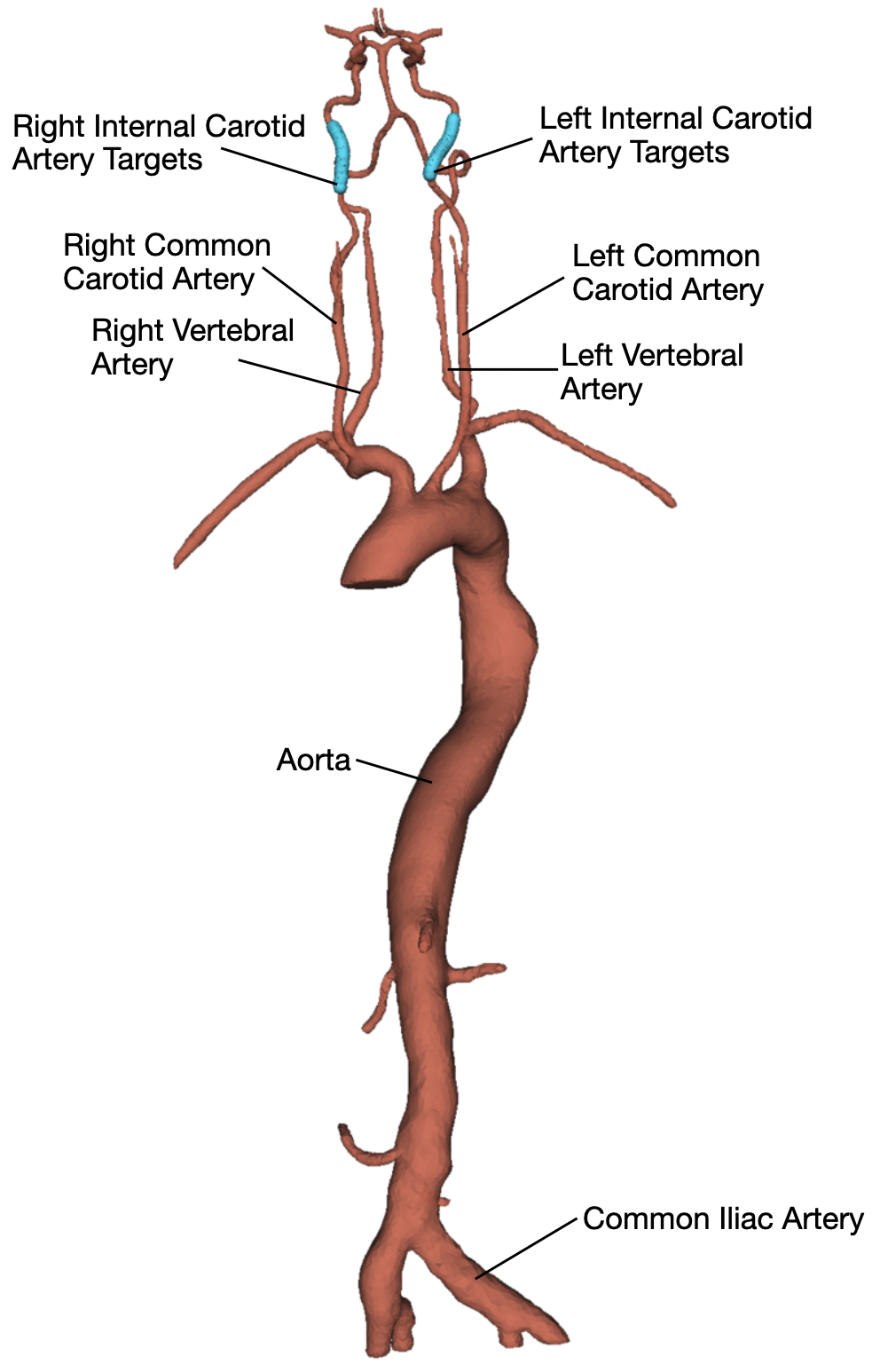}
                 \caption{ Example targets.}
                 \label{fig:MT_env}
            \end{subfigure}
            \begin{subfigure}[b]{0.38\textwidth}
                \includegraphics[width=\textwidth]{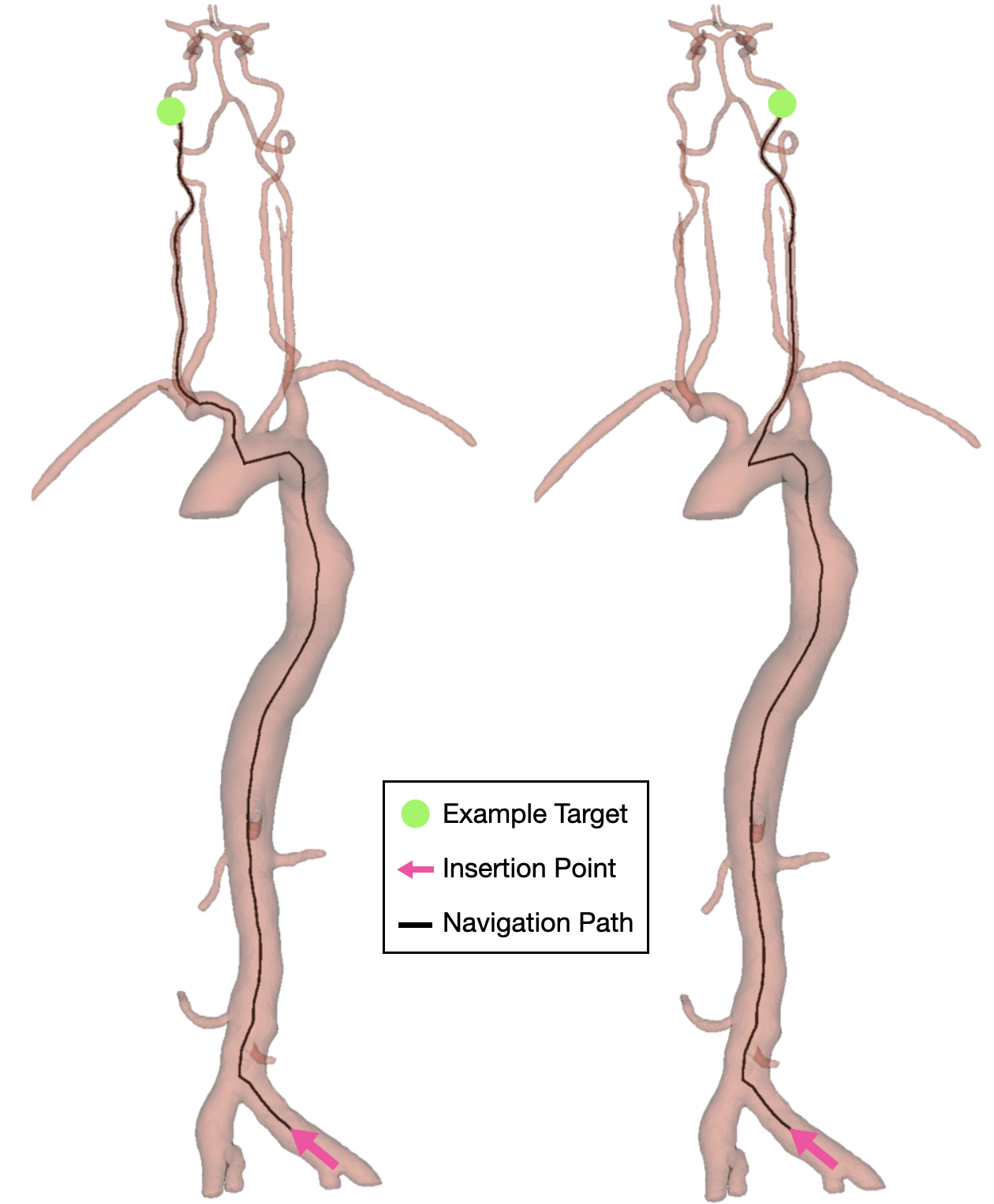}
                \caption{ Navigation path.}
                \label{fig:nav_path}
            \end{subfigure}
            \caption{ MT environment \textbf{a} used in simulations, with anatomy labelled and all possible targets, \textbf{b} with insertion point and example navigation path to target in each common carotid artery.}
            \label{fig:results}
        \end{figure}

        In MT, a guidewire is typically used to navigate a guide catheter to the ICA. An `access catheter' – one with a distal shape to allow access into vessel branches – is placed within the guide catheter and taken ahead of the guide catheter tip during navigation. Once the access catheter is within the ICA, the guide catheter is advanced to make a stable platform. At this point, the guidewire and access catheter are retracted, and a micro-guidewire within a micro-catheter are together passed through the stable guide catheter and navigated to the target site. The navigation task presented in this paper simulates this complete navigation of guidewire and access catheter. A target location, randomly sampled from 30 centreline points within the right internal carotid artery (RICA) or left internal carotid artery (LICA), is chosen to be navigated from an insertion point in the common iliac artery. 20 targets in each branch were selected for training, and the remaining 10 holdout target locations were used for evaluation. For each navigation attempt, the train or test target location is changed. Fig.~\ref{fig:MT_env} shows the simulation environment, with anatomy labelled, and the possible targets. Fig.~\ref{fig:nav_path} shows the insertion point (standard across all runs) and an example navigation path to a target within each carotid artery.

    \subsection{Simulation environment}\label{environment}

        The \textit{in silico} environment for the navigation task builds upon prior contributions from \cite{Jackson2023, Karstensen2023}, both of which use the Simulation Open Framework Architecture (SOFA), along with the modified BeamAdapter plugin, to create a comprehensive simulation-based training and evaluation environment \cite{Faure2012, Duriez2006}. SOFA allowed replication of vascular structures derived from two computed tomography angiography (CTA) scans. The first scan encompassed the abdominal and thoracic regions, including the femoral arteries, descending aorta, and the aortic arch. The second CTA scan encompassed the aortic arch and extended to the cerebral vessels, including the carotid and vertebral arteries in the neck. These scans were subsequently processed into surface meshes, each consisting of approximately 850 vertices, which were then manually merged to create a cohesive and continuous model using Blender \cite{Blender2018}.

        Integrating catheters (characterized by 160 vertices with Young's modulus of 47\,\unit{\mega\pascal}) and guidewires (comprising 120 vertices with Young's modulus of 43\,\unit{\mega\pascal}) within the \textit{in silico} environment were facilitated by the BeamAdapter plugin developed for SOFA. This plugin was used to model the Penumbra Neuron MAX 088 guide catheter (Penumbra, California, USA) and the Terumo 0.035 guidewire (Terumo, Tokyo, Japan) to generate the initial topology and mechanical properties needed for the simulation. 

        The simulation assumed rigid vessel walls with an empty lumen. The simulation's fidelity to real-world guidewire behaviour was achieved through iterative fine-tuning of parameters such as the friction between the guidewire and vessel wall and the guidewire's stiffness, as demonstrated by \cite{Karstensen2023} and \cite{Karstensen2022}, the latter of which has shown promising translation from \textit{in silico} to an \textit{ex vivo} porcine liver. A qualitative survey shows that six experienced (greater than five years clinical experience) interventional neuroradiologists (UK consultant grade; US attending equivalent), two novice (less than three years clinical experience) interventional neuroradiologists and one interventional radiologist (UK specialist trainee grade; US fellow equivalent) agreed that the vessel anatomy was accurate, agreed the guide catheter performed realistically, and were indifferent to the realism of the guidewire \cite{Jackson2023}. 
        
        Input parameters for the simulation, including guidewire rotation and translation speed, were applied at the proximal end of the catheter and guidewire. The output of the simulation was the catheter and guidewire position as a series of coordinate points directly extracted from the finite-element model. The rotational speed was constrained to a maximum of 180\,\unit{\degree\per\second}, while the translational speed was capped at 40\,\unit{\mm\per\second}. Feedback during the navigation was given as two-dimensional $(x',z')$ tracking coordinates of the device, and no feedback about the vessel geometry was given. Experiments were performed using solely guidewire-tracking (single-tracking) and combined guidewire and catheter-tracking (dual-tracking) methods.

    \subsection{Demonstrator data collection}

        The navigation task was performed \textit{in silico} on ten random targets in each carotid artery for demonstrator data collection. Navigation of the targets from the same insertion point in Fig.~\ref{fig:nav_path} was performed using inputs from a keyboard. The action (device rotation and translation), device tracking data, and target location were collected at each simulation step. Data was split based on if the target branch was on the left or right side. An example navigation path of guidewire and catheter can be seen in Fig.~\ref{fig:Nav_path_demo}.

    \subsection{IRL algorithm}

        IRL was used to determine a suitable reward function from the collected demonstrator data. The maximum entropy IRL (MaxEnt IRL) algorithm was employed to learn the reward functions based on expert demonstrations, specifically for navigating through different branches of vasculature \cite{Ziebart2008}. The algorithm's objective is twofold: it seeks to replicate the observed behavior of experts and ensures that the learned reward functions capture a distribution of behavior with maximum entropy. To achieve this, MaxEnt IRL involves an iterative process of updating reward functions to maximize the entropy of the expected policy while maintaining consistency with observed expert behavior. The reward functions are initialized randomly, with the same random seed across experiments, and updated using gradient ascent to minimize the entropy of the policy.
        
        For each branch, the reward function is updated using the gradient ascent step in Eq.~\ref{eq:RF}, where $R_i$ is the reward function for branch $i$, $\nabla R_i$ is the gradient vector, and $\eta$ is the learning rate. 
        
        \begin{equation}
            R_i \leftarrow R_i + \eta \nabla R_i 
            \label{eq:RF}
        \end{equation}
        
        The MaxEnt IRL objective is expressed as in Eq.~\ref{eq:MaxEnt}, for a policy $\pi\left ( a\mid s \right )$, which is the probability distribution over actions given a particular state.
        
        \begin{equation}
            \textup{Maximize  } \mathbb{E}_\pi \left [  -\sum_a \pi\left ( a\mid s \right ) \log \pi \left ( a\mid s \right ) \right ]
            \label{eq:MaxEnt}
        \end{equation}
        
        The Boltzmann policy calculation for action $a$ given state $s$ is defined as Eq.~\ref{eq:Boltzman}.
        
        \begin{equation}
            \pi \left ( a\mid s \right ) = \frac{e^{R_i \left ( s,a \right )}}{\sum _{a^i}e^{R_i \left ( s,{a}' \right )}}
            \label{eq:Boltzman}
        \end{equation}
        
        A feedforward neural network with four fully connected layers trained the model using expert trajectories, optimising the policy through $1 \times 10^6$ iterations. A separate model was trained for each set of targets in each branch. At the start of a new navigation task, the correct model based on the target branch was obtained, and at each step, a reward value was returned based on the current observation given to the model.

    \subsection{Controller architecture, reward functions and training}

        In this study, all RL models were trained using a SAC controller, adapted to facilitate two-device tracking and accommodate various reward functions from the architecture developed in previous work by Karstensen et al. \cite{Karstensen2023}. The architecture consists of a Long Short-Term Memory (LSTM) layer, which functions as an observation embedder, enabling the learning of a trajectory-dependent state representation. The subsequent feedforward layers are responsible for learning the control of the guidewire. The LSTM-based observation embedder is updated exclusively using the q1-network. In this architecture, the controller receives an observation as input, and the Gaussian policy network generates parameters, specifically mean ($\mu$) and standard deviation ($\sigma$), of a normal distribution for the following action. During training, actions are sampled from this normal distribution, whereas for evaluation, $\mu$ is directly employed as the action, rendering the behaviour deterministic. Training was completed on a NVIDIA DGX A100 node with 8 GPUs (Santa Clara, California, USA).

        The action was defined as the output of the Gaussian policy network representing the catheters' and the guidewires' rotation and translation speed. Three points on the instrument tip described the instrument's position, denoted as $(x', z'){i=1,2,3}$, with $(x', z'){1}$ coinciding with the instrument tip. Additionally, the target position was specified by the current target's $(x', z')$-coordinates. The observation comprised the current and previous guidewire and catheter positions, the target position, and the previous action taken from the last to the current position.
    
        The controller training procedure performed the navigation task for $1\times10^{7}$ exploration steps, defined as a single control loop cycle during the exploration phase. Each navigation task was considered an episode and was deemed complete once the target was reached within a 5 mm threshold. To ensure computational efficiency, we introduced a timeout after 400 exploration steps (equivalent to approximately 53\,s) without reaching the target. The control frequency was set to 7.5~Hz.

        Three types of reward functions were used in this study to determine whether deriving a reward function from demonstrator data provides any benefit for autonomous endovascular navigation. Each reward function is calculated in every simulation step from the environment state based on the agent's actions. Therefore, actions will lead to different reward quantities across reward functions and, hence, a variation in the final model. 
        
        The current state-of-the-art algorithm for autonomous navigation in endovascular interventions is devised by \cite{Karstensen2023}. This algorithm uses a dense reward function, $R_{1}$, which does not use IRL and is shown in Eq.~\ref{eq:R1}. Here, \textit{pathlength} is defined as the distance between the guidewire tip and the target along the centerlines of the arteries, with $\Delta\text{pathlength}$ representing the change in pathlength at time $t$ from the previous step at time $t=-1$.

        \begin{equation}
            R_{1} = -0.005 - 0.001\cdot\Delta\text{pathlength}+\begin{cases}1.0 & \text{if target reached}\\0 & \text{else}\end{cases}
            \label{eq:R1}
        \end{equation}

        The second reward function $R_{2}$ was derived using IRL, as seen in Eq.~\ref{eq:R2}, where $R_{RICA}$ and $R_{LICA}$ are equal to $R_{i}$ in Eq.\ref{eq:RF} calculated for the right and left carotid arteries, respectively.

        \begin{equation}
            R_{2} = \begin{cases}
                R_{RICA} & \text{if target in RICA}\\
                R_{LICA} & \text{if target in LICA}
            \end{cases}
            \label{eq:R2}
        \end{equation}

        The third reward function $R_{3}$ was obtained through reward shaping using a combination of the dense reward function $R_{1}$ and the IRL-derived reward function $R_{2}$, as shown in Eq.~\ref{eq:R3}. $\alpha$ is a scaling factor used to provide the correct ratio of $R_{1}$ and $R_{2}$. A scaling factor of 0.001 was used in this study.

        \begin{equation}
            R_{3} = R_{1} +  \alpha R_{2}
            \label{eq:R3}
        \end{equation}

        Evaluations were conducted every $2.5 \times 10^5$ exploration steps for 100 episodes. Ten targets in each branch unused during training were utilized for evaluation. Each evaluation step records the success rate, procedural times, and path ratio. Comparison is performed between the highest success rate of each model, and the corresponding path ratio and procedure times at this evaluation step. The \textit{Success Rate} is the percentage of evaluation episodes in which the controller successfully reaches the target; \textit{Path Ratio} is a measure of the remaining distance to the target point in unsuccessful episodes, calculated by dividing the remaining distance by the initial distance; and \textit{Procedure Time} is the time taken from the start of navigation to the target location for successful episodes. \textit{Exploration Steps} is the number of training steps taken to reach the point at which the results are provided. Comparative statistical analyses were conducted using two-tailed paired Student's t-tests, with a predetermined significance threshold set at p = 0.05.

\section{Results}\label{sec3}

    \subsection{Single vs dual device tracking}

\begin{figure}[t!]
            \centering
            \begin{subfigure}{8cm}
                 \includegraphics[width=\hsize]{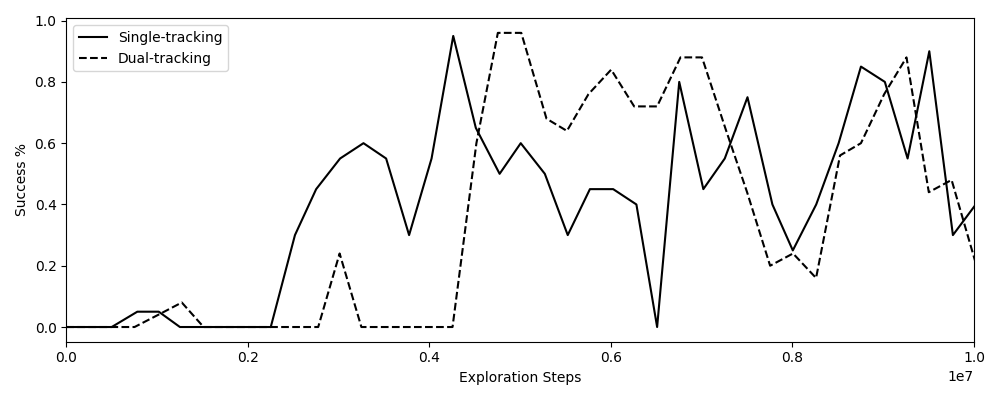}
                 \caption{Success rate (\%)}
                 \label{fig:SRvsStep_track}
            \end{subfigure}
            \hfill
            \begin{subfigure}{8cm}
                \includegraphics[width=\hsize]{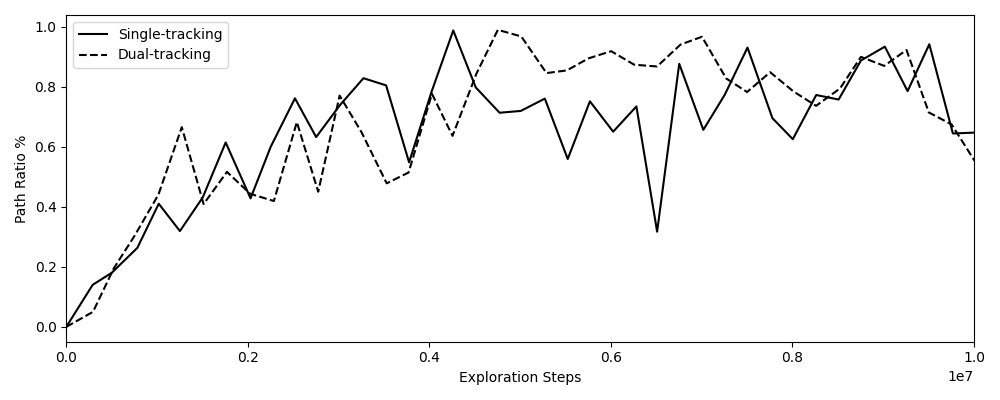}
                \caption{Path ratio (\%)}
                \label{fig:TLvsStep_track}
            \end{subfigure}
            \caption{ \textbf{a} Success rate (\%), \textbf{b} Path ratio  (\%) during training for single vs dual device tracking.}
            \label{fig:results_track}
        \end{figure}

\begin{figure}[t!]
            \centering
            \begin{subfigure}{2.65cm}
                 \includegraphics[width=\hsize]{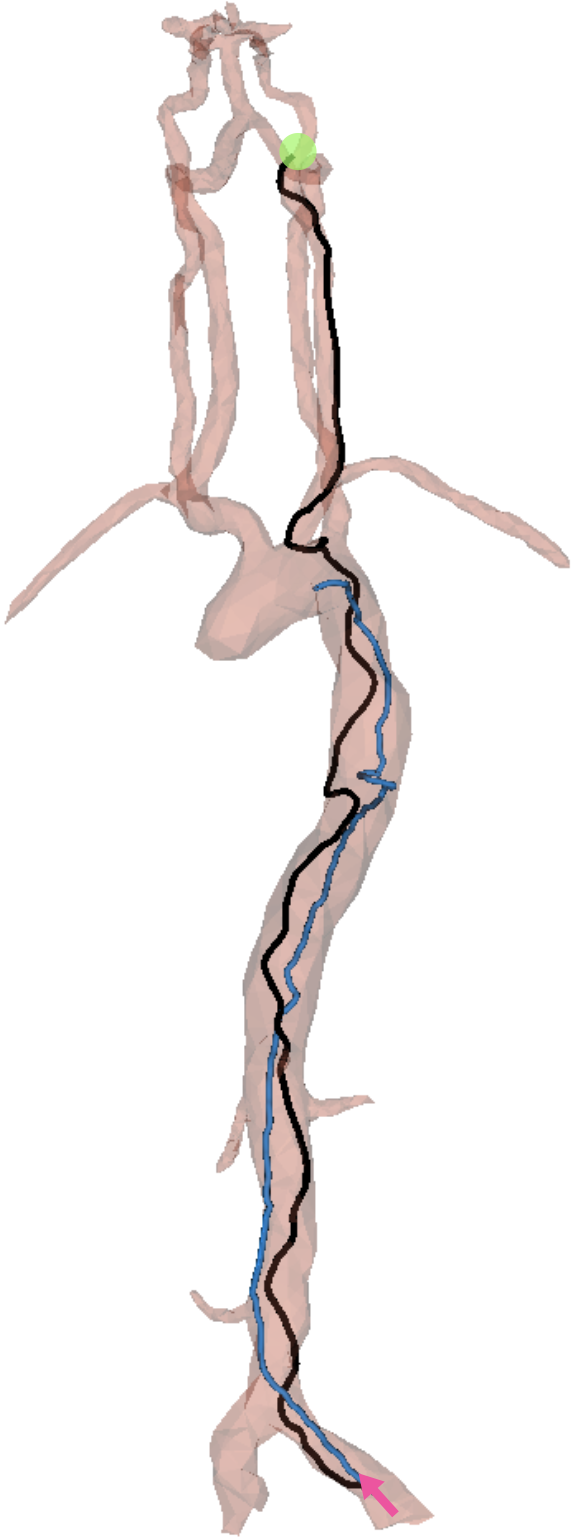}
                 \caption{Demonstrator}
                 \label{fig:Nav_path_demo}
            \end{subfigure}
            \begin{subfigure}{3.5cm}
                 \includegraphics[width=\hsize]{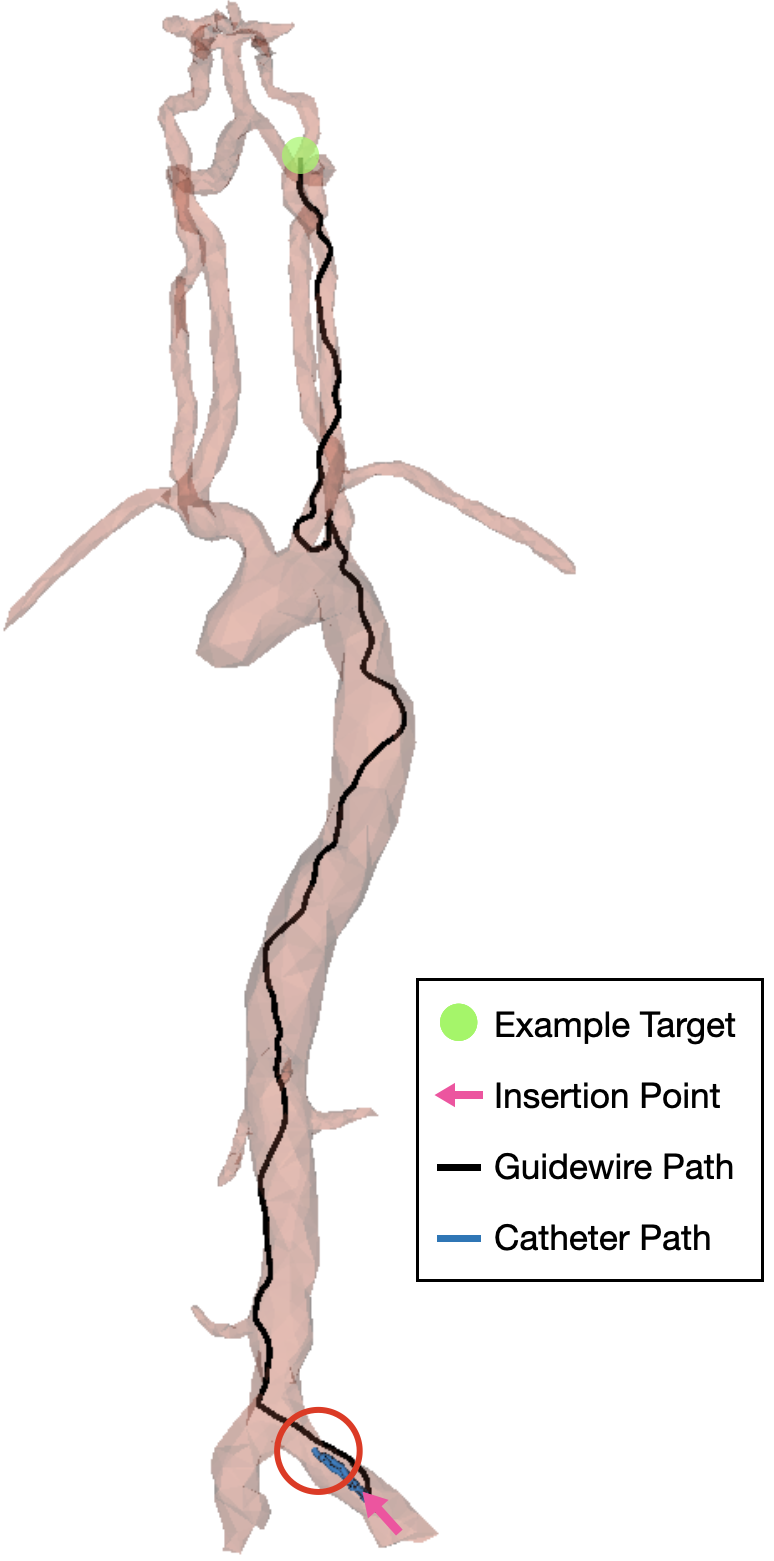}
                 \caption{Single-tracking}
                 \label{fig:Nav_path_single}
            \end{subfigure}
            \begin{subfigure}{2.73cm}
                \includegraphics[width=\hsize]{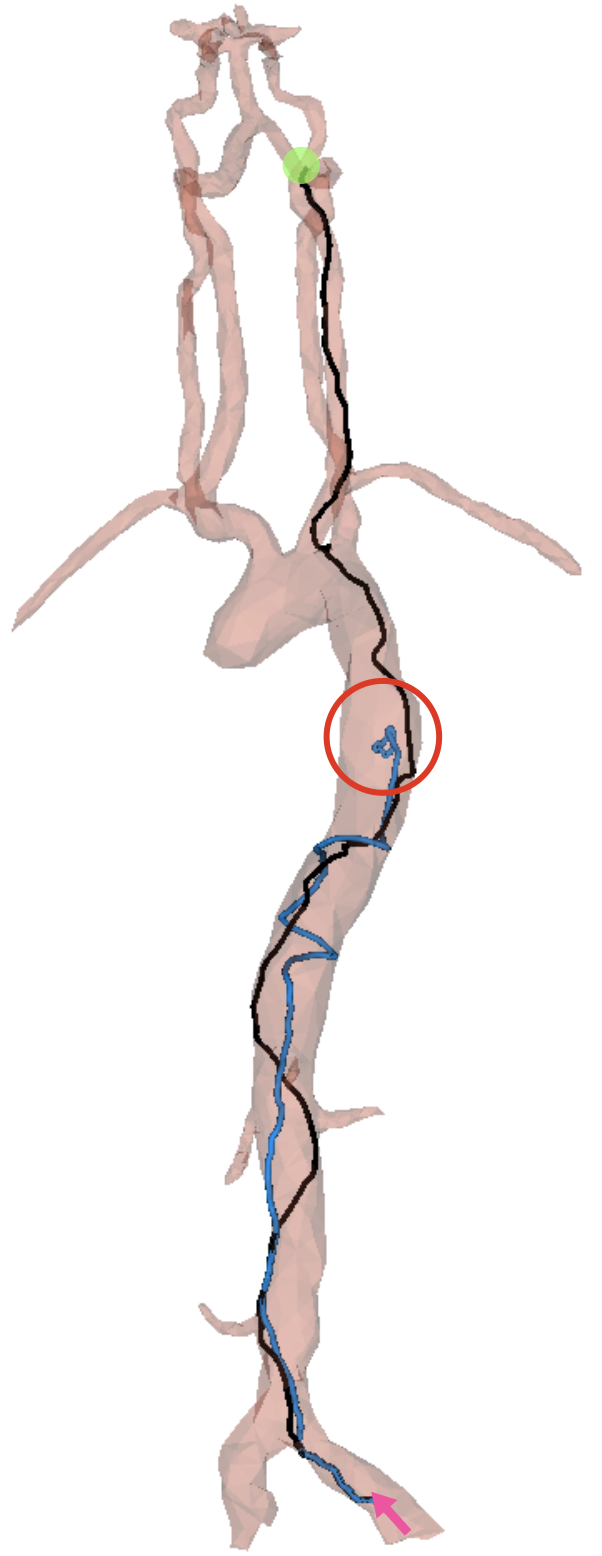}
                \caption{Dual-tracking}
                \label{fig:Nav_path_dual}
            \end{subfigure}
            \begin{subfigure}{2.7cm}
                \includegraphics[width=\hsize]{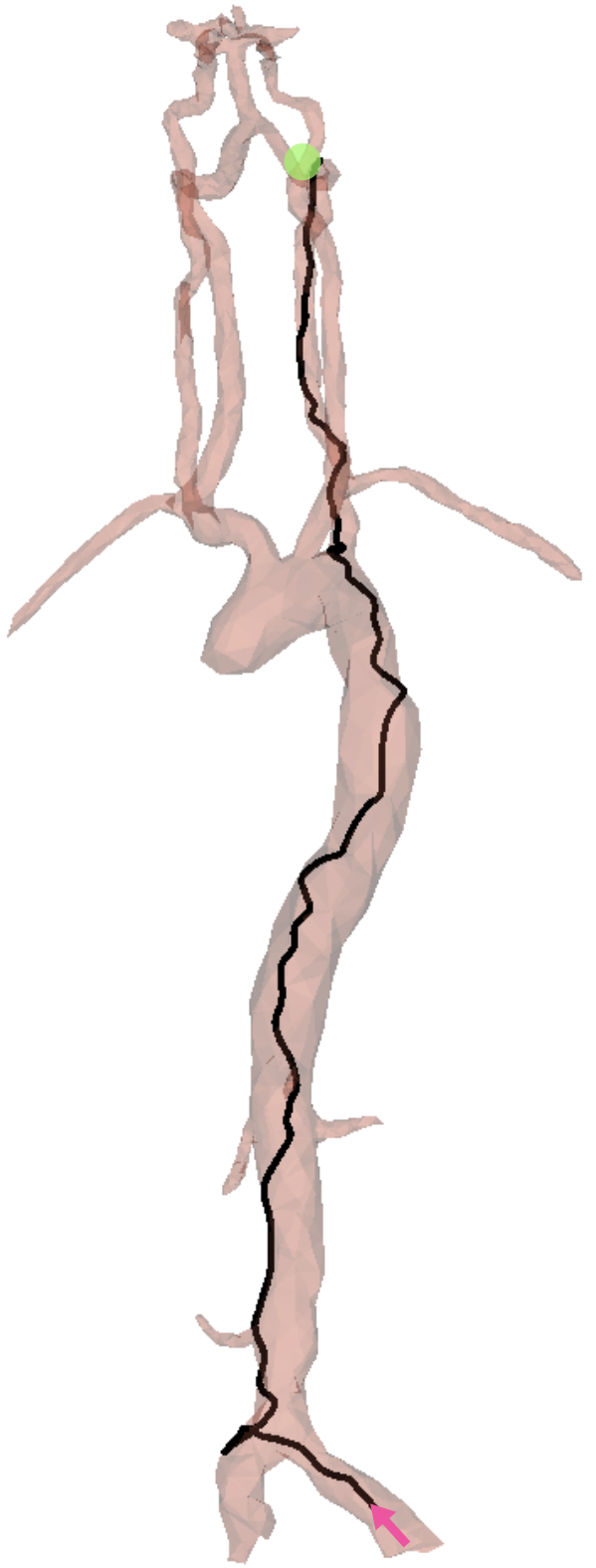}
                \caption{IRL}
                \label{fig:Nav_path_irl}
            \end{subfigure}
            \caption{ Trajectories of catheter and guidewire tip for \textbf{a} demonstrator data, \textbf{b} single device tracking (with final catheter position highlighted), \textbf{c} dual device tracking (with final catheter position highlighted), and \textbf{d} IRL.}
            \label{fig:results_nav_path}
        \end{figure}

        Results of an investigation into the difference in success rate, procedure times and path ratio during training between tracking solely the guidewire and tracking both the guidewire and catheter are shown in Table~\ref{tab:tracking}. Fig.~\ref{fig:results_track} shows the success rate and path ratio during training. Single-tracking and dual-tracking reach similar maximum success rates of 95\% and 96\% respectively ($p= 0.741$), with single-tracking having a lower procedure time of 22.5\,s compared to 24.9\,s for dual-tracking ($p=0.047$). Fig.~\ref{fig:results_nav_path} compares the guidewire and catheter trajectories from insertion to a target in the LICA of single and dual device tracking. The final tip position of catheters for these two models is highlighted for comparison. For single-tracking, the catheter is utilized sparingly and is not navigated out of the common iliac artery. In contrast, during dual-tracking navigation, the catheter is used throughout the navigation task and is taken up the aortic arch to provide stability. Dual-tracking navigation, therefore, mimics neurointerventional radiology experts, whereas single-tracking navigation does not.

        \begin{table}[t!]
            \centering
            \caption{Results of device tracking training}
            \label{tab:tracking}
            \begin{tabular}{l|l|l|l|l}
                \textbf{\begin{tabular}[c]{@{}l@{}}Tracking \\ Method\end{tabular}} & \textbf{\begin{tabular}[c]{@{}l@{}}Success \\ Rate (\%)\end{tabular}} & \textbf{\begin{tabular}[c]{@{}l@{}}Procedure \\ Time (s)\end{tabular}} & \textbf{\begin{tabular}[c]{@{}l@{}}Path \\ Ratio (\%)\end{tabular}} & \textbf{\begin{tabular}[c]{@{}l@{}}Exploration \\ Steps\end{tabular}} \\ \hline
                Single-tracking & 95 & 22.5 & 98.7 & $4.26 \times 10^6$\\ \hline
                Dual-tracking   & 96 & 24.9 & 98.9 & $4.75 \times 10^6$
            \end{tabular}
        \end{table}

    \subsection{IRL and reward shaping}
    
\begin{figure}[t!]
            \centering
            \begin{subfigure}{8cm}
                 \includegraphics[width=\hsize]{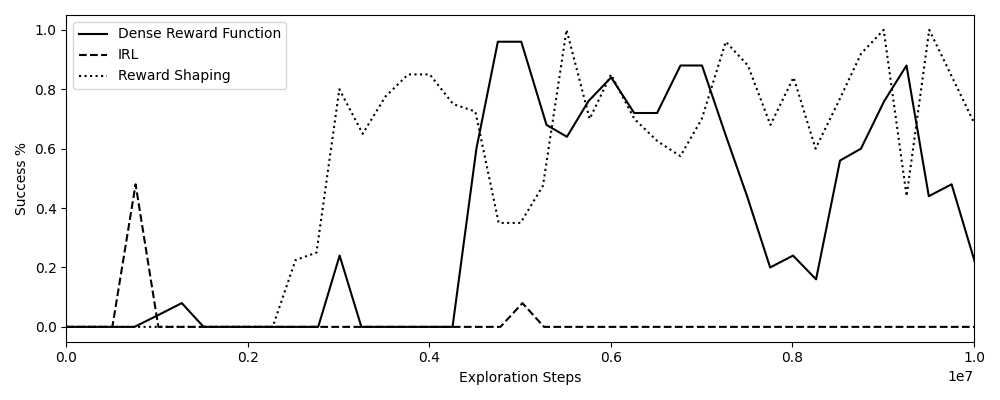}
                 \caption{Success rate}
                 \label{fig:SRvsStep_IRL}
            \end{subfigure}
            \hfill
            \begin{subfigure}{8cm}
                \includegraphics[width=\hsize]{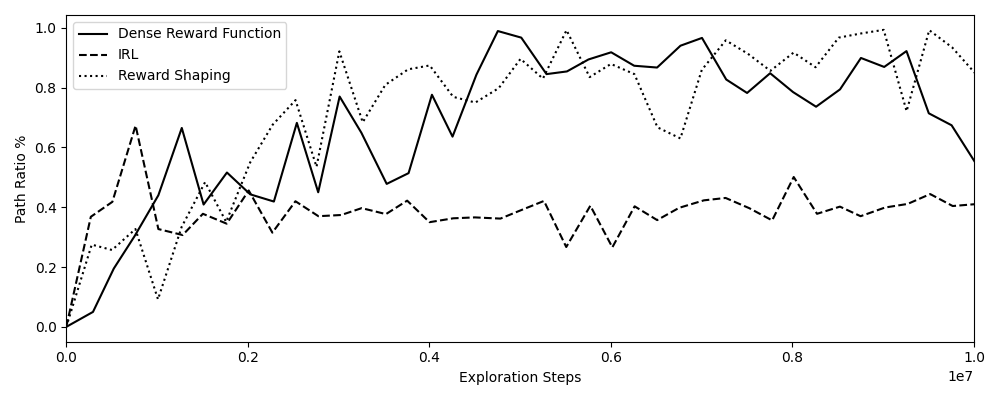}
                \caption{Path ratio}
                \label{fig:TLvsStep_IRL}
            \end{subfigure}
            \caption{ \textbf{a} Success rate, \textbf{b} Path ratio during training for a dense reward function, IRL-derived reward function, and reward shaping function.}
            \label{fig:results_IRL}
        \end{figure}

        Table~\ref{tab:results_irl} shows the final success rate, procedure time and path ratio for a dense reward function alone, an IRL-derived reward function alone, and a reward function obtained through reward shaping (which combines a dense reward function and an IRL-derived reward function). All experiments are performed using dual-tracking. Fig.~\ref{fig:results_IRL} shows the success rate and path ratio during training. Reward shaping provides the highest success rate of 100\%, followed by 96\% for the dense reward function ($p=0.045$). Reward shaping also has the lowest procedure time out of the examined reward functions at 22.6\,s, compared to 24.9\,s for the dense reward function ($p=0.0001$). Fig.~\ref{fig:Nav_path_irl} shows an example navigation path for guidewire and catheter of the IRL-derived reward function at $0.76 \times 10^6$ exploration steps, where it can be seen that the low success rate of 48\% can be attributed to catheterization of the left vertebral artery, rather than the right carotid artery.
    
        \begin{table}[t!]
            \centering
            \caption{Results of reward function training}
            \label{tab:results_irl}
            \begin{tabular}{l|l|l|l|l}
                \textbf{\begin{tabular}[c]{@{}l@{}}Tracking \\ Method\end{tabular}} & \textbf{\begin{tabular}[c]{@{}l@{}}Success \\ Rate (\%)\end{tabular}} & \textbf{\begin{tabular}[c]{@{}l@{}}Procedure \\ Time (s)\end{tabular}} & \textbf{\begin{tabular}[c]{@{}l@{}}Path \\ Ratio (\%)\end{tabular}} & \textbf{\begin{tabular}[c]{@{}l@{}}Exploration \\ Steps\end{tabular}} \\ \hline
                Dense           & 96  & 24.9 & 98.9 & $4.75 \times 10^6$ \\ \hline
                IRL             & 48   & 59.2 & 67.3 & $0.76 \times 10^6$ \\ \hline
                Reward Shaping  & 100 & 22.6 & 100 & $5.51 \times 10^6$
            \end{tabular}
        \end{table}

\section{Discussion}\label{sec4}

    This study was the first to perform a navigation task that simulated the MT intervention by autonomously navigating both guidewire and catheter from an insertion point in the femoral artery to a target in the carotid artery. Furthermore, our study introduced for the first time dual device for autonomous endovascular navigation, providing information about catheters and guidewire during RL training. This study was also the first to investigate reward functions in autonomous endovascular navigation using IRL to identify a specific reward function from demonstrator data for RL training.

    There have been no studies applying RL in any form to any task related to MT \cite{Robertshaw2023}. Our study demonstrated proof-of-concept of \textit{in silico} autonomous navigation of instruments in MT. The implication is that if further validation studies demonstrate benefit, our approach can plausibly increase patient accessibility to MT while increasing intervention safety and speed.

    \subsection{Single vs dual tracking}

        Both models showed an increase in path ratio and success rate over training. Results showed that single and dual-tracking models reach a similarly high success rate after training, but the single-tracking method records quicker procedure times. However, investigation into the dual-tracking models shows that the catheter is utilized from an early stage during navigation, with the movements performed similarly to those of an experienced neurointerventional radiologist. The catheter has a higher stiffness than the guidewire, and hence, in the dual-tracking model, the catheter is taken up the aortic arch to provide stability as the guidewire is navigated into the branching point of the common carotid arteries. For this reason, dual-tracking was used in all reward function experiments, as it better replicates the actions of an experienced operator, with no trade-off for success rate.
        
        It is noteworthy that in this experiment, a single static mesh and a relatively simple navigation task were used. It is plausible that in a more challenging navigation task, for example, where the anatomy is more tortuous, as seen in old and hypertensive patients, the dynamic use of two instruments (dual-tracking with guidewire and catheter) would enable a higher success rate.

    \subsection{Reward function}

        Results showed that reward shaping using a combined learned IRL and a dense reward function gives a higher overall success rate, and a lower procedure time than the state-of-the-art \cite{Karstensen2023}. The model trained using reward shaping can use demonstrator data through IRL to navigate toward the target, while the dense reward function provides an incentive for reaching and moving towards the target while promoting this in the lowest number of steps possible. 
        
        Training using only an IRL-derived reward function did not reach the same success rates as the other models, although it can be seen that the model was able to navigate towards the target. However, the low success rate comes from wrong branch catheterization, which could be caused by the IRL-derived reward function not taking into account the vessel walls and, therefore, receiving a high reward signal for a small Euclidean distance in the coordinate system, while the \textit{pathlength} is still significant. For anatomical clarity, the distal vertebral artery and the ipsilateral internal carotid artery are approximately 10\,\unit{\mm} from each other yet both arteries are in continuity with the aortic arch allowing the guidewire to move seamlessly to either location. These issues were overcome through reward shaping, as by providing a negative reward for moving away from the target via \textit{pathlength}, the success rate can be increased dramatically. 

        A model with a high path ratio is still valid for all experiments performed within this study. Failures where the path ratio was above 80\% usually meant that the correct branch was catheterized, however, the guidewire did not reach the target point. Despite this, the guidewire would be in the correct region of the vessel and be helpful for the operator and subsequent steps in MT. After all, in MT the initial aim simulated by our experiments is to navigate to this approximate area within the ICA with a standard guidewire facilitated by an access catheter. Thereafter, a guide catheter is advanced to this approximate area, and then a micro-guidewire and microcatheter is used. The exact location of the initial navigation within the ICA is often less important than simply having a guidewire in the ICA that is stable enough to allow the subsequent endovascular steps.

    \subsection{Limitations}

        While the results of this investigation towards an autonomous MT navigation system fall under a technology readiness level (TRL) of 3 \cite{Mankins1995}, this proof-of-concept work showed that IRL can be leveraged effectively as part of reward shaping to deduce a suitable reward function for RL training, there are limitations to the methodology employed in terms of utility. \textit{In vitro} (e.g., phantom) and clinical validation steps would be required to progress the TRLs.
        
        First, demonstrator data was obtained using a simple keypad controller; therefore, to improve the reward function's suitability, data could be collected to accurately reflect the actions performed by an experienced neurointerventional radiologist. Furthermore, 20 demonstration trajectories were used for IRL, and a higher sample size may provide a more generalized reward function. Additionally, the reward function could be altered to give consideration to safety during the intervention. By providing a negative reward for vessel wall contact forces over a threshold, the puncturing of a vessel wall causing a perforation can be prevented.

        Second, while different targets were used for training and testing, the same mesh was used throughout the experiments. Future \textit{in silico} validation work should apply these algorithms to a range of patient anatomy (a dataset with multiple training and hold-out testing unique patient anatomy) to increase generalizability, improving the effectiveness during future \textit{in vitro} experiments.

        Third, simulations in this study were performed using tracking coordinates of the guidewire and catheter tip. An alternative to this would be to use an image-based tracking system, which may have more clinical relevance, as the autonomous system could use fluoroscopic imaging that already takes place during MT interventions. For future \textit{in vitro} experiments, further work will investigate the effectiveness of the techniques demonstrated in this study to imaged-based tracking inputs and those captured by device tracking.

        Fourth, recent clinical trials of robotic diagnostic cerebral angiography recommend that procedural times, success rates, fluoroscopy times, and radiation doses be compared with the traditional manual approach \cite{Beaman2023}. The measurement of fluoroscopy times and radiation doses is outside the scope of this \textit{in silico} study. However, success rate, procedural times, and path ratio were recorded at each evaluation step. Future \textit{in vitro} experiments could measure fluoroscopy times and radiation doses.

        Our study successfully addresses the task of simulating the complete navigation of guidewire and guide catheter in MT, which takes place from the common iliac artery to either the LICA or RICA where the guide catheter is in a stable position. However, a complete MT procedure typically involves a subsequent navigation step from the LICA or RICA to the MCA. Such a subsequent navigation to the MCA requires different instrumentation (often microcatheters and microwires) beyond the scope of our investigation. Therefore, while this study provides valuable insights into a specific aspect of the MT procedure involving several steps where there is challenging anatomy, it does not represent a comprehensive examination of the entire MT navigation procedure.

\section{Conclusion}\label{sec5}
    
    We showed the feasibility of autonomous guidewire navigation throughout the MT vasculature using IRL with demonstrator data for the first time. We established a comprehensive simulation-based training environment and, for the first time, compared models with different reward functions for autonomous endovascular navigation by utilizing SOFA and SAC. Reward shaping emerged superior, with higher success rates and lower procedure times than the state-of-the-art. Single versus dual device tracking revealed that both methods achieved high success rates, but dual-tracking utilized both devices, effectively mimicking experienced neurointerventional radiologists. We have highlighted several opportunities for future research beyond our proof-of-concept experiments, in particular relating to improving performance. Another opportunity is to investigate the final stage of navigation which is typically from the distal aspect of the cervical ICA to the M1 segment of the MCA using microcatheter and microwire. Additionally, researchers may wish to apply our approach to other non-MT endovascular systems. In summary, this work offers promising avenues for improved procedural accessibility and precision of \textit{in silico} autonomous endovascular navigation tasks. For MT, the work plausibly lays the foundation for potentially transformative patient care.

\section*{Funding}
Partial financial support was received from the Wellcome/ EPSRC Centre for Medical Engineering [WT203148/Z/16/Z] and from the Engineering and Physical Sciences Research Council Doctoral Training Partnership (EPSRC DTP) under Grant Agreement No EP/R513064/1. For the purpose of Open Access, the Author has applied a CC BY public copyright license to any Author Accepted Manuscript version arising from this submission.

\bibliographystyle{unsrt}  
\bibliography{references}

\begin{thebibliography}{10}

\bibitem{Townsend2016}
Nick Townsend, Lauren Wilson, Prachi Bhatnagar, Kremlin Wickramasinghe, Mike Rayner, and Melanie Nichols.
\newblock Cardiovascular disease in europe: Epidemiological update 2016.
\newblock {\em European Heart Journal}, 37:3232--3245, 11 2016.

\bibitem{Goyal2016}
Mayank Goyal, Bijoy~K. Menon, Wim H.~Van Zwam, Diederik~W.J. Dippel, Peter~J. Mitchell, Andrew~M. Demchuk, Antoni Dávalos, Charles~B.L.M. Majoie, Aad Van~Der Lugt, Maria A.~De Miquel, Geoffrey~A. Donnan, Yvo~B.W.E.M. Roos, Alain Bonafe, Reza Jahan, Hans~Christoph Diener, Lucie A. Van~Den Berg, Elad~I. Levy, Olvert~A. Berkhemer, Vitor~M. Pereira, Jeremy Rempel, Mònica Millán, Stephen~M. Davis, Daniel Roy, John Thornton, Luis~San Román, Marc Ribó, Debbie Beumer, Bruce Stouch, Scott Brown, Bruce~C.V. Campbell, Robert J.~Van Oostenbrugge, Jeffrey~L. Saver, Michael~D. Hill, and Tudor~G. Jovin.
\newblock Endovascular thrombectomy after large-vessel ischaemic stroke: A meta-analysis of individual patient data from five randomised trials.
\newblock {\em The Lancet}, 387:1723--1731, 4 2016.

\bibitem{Vidale2017}
Simone Vidale and Elio Agostoni.
\newblock Endovascular treatment of ischemic stroke: An updated meta-analysis of efficacy and safety.
\newblock {\em Vascular and Endovascular Surgery}, 51:215--219, 5 2017.

\bibitem{Rha2007}
Joung~Ho Rha and Jeffrey~L. Saver.
\newblock The impact of recanalization on ischemic stroke outcome: A meta-analysis.
\newblock {\em Stroke}, 38:967--973, 3 2007.

\bibitem{Berkhemer2015}
Olvert~A. Berkhemer, Puck~S.S. Fransen, Debbie Beumer, Lucie~A. van~den Berg, Hester~F. Lingsma, Albert~J. Yoo, Wouter~J. Schonewille, Jan~Albert Vos, Paul~J. Nederkoorn, Marieke~J.H. Wermer, Marianne~A.A. van Walderveen, Julie Staals, Jeannette Hofmeijer, Jacques~A. van Oostayen, Geert J.~Lycklama a~Nijeholt, Jelis Boiten, Patrick~A. Brouwer, Bart~J. Emmer, Sebastiaan~F. de~Bruijn, Lukas~C. van Dijk, L.~Jaap Kappelle, Rob~H. Lo, Ewoud~J. van Dijk, Joost de~Vries, Paul~L.M. de~Kort, Willem Jan~J. van Rooij, Jan~S.P. van~den Berg, Boudewijn~A.A.M. van Hasselt, Leo~A.M. Aerden, René~J. Dallinga, Marieke~C. Visser, Joseph~C.J. Bot, Patrick~C. Vroomen, Omid Eshghi, Tobien~H.C.M.L. Schreuder, Roel~J.J. Heijboer, Koos Keizer, Alexander~V. Tielbeek, Heleen~M. den Hertog, Dick~G. Gerrits, Renske~M. van~den Berg-Vos, Giorgos~B. Karas, Ewout~W. Steyerberg, H.~Zwenneke Flach, Henk~A. Marquering, Marieke~E.S. Sprengers, Sjoerd~F.M. Jenniskens, Ludo~F.M. Beenen, Rene van~den Berg, Peter~J. Koudstaal, Wim~H. van Zwam,
  Yvo~B.W.E.M. Roos, Aad van~der Lugt, Robert~J. van Oostenbrugge, Charles~B.L.M. Majoie, and Diederik~W.J. Dippel.
\newblock A randomized trial of intraarterial treatment for acute ischemic stroke.
\newblock {\em New England Journal of Medicine}, 372:11--20, 1 2015.

\bibitem{Saver2016}
Jeffrey~L. Saver, Mayank Goyal, Aad Van~Der Lugt, Bijoy~K. Menon, Charles~B.L.M. Majoie, Diederik~W. Dippel, Bruce~C. Campbell, Raul~G. Nogueira, Andrew~M. Demchuk, Alejandro Tomasello, Pere Cardona, Thomas~G. Devlin, Donald~F. Frei, Richard Du Mesnil~De Rochemont, Olvert~A. Berkhemer, Tudor~G. Jovin, Adnan~H. Siddiqui, Wim H.~Van Zwam, Stephen~M. Davis, Carlos Castaño, Biggya~L. Sapkota, Puck~S. Fransen, Carlos Molina, Robert J.~Van Oostenbrugge, Ángel Chamorro, Hester Lingsma, Frank~L. Silver, Geoffrey~A. Donnan, Ashfaq Shuaib, Scott Brown, Bruce Stouch, Peter~J. Mitchell, Antoni Davalos, Yvo~B.W.E.M. Roos, and Michael~D. Hill.
\newblock Time to treatment with endovascular thrombectomy and outcomes from ischemic stroke: Ameta-analysis.
\newblock {\em JAMA - Journal of the American Medical Association}, 316:1279--1288, 9 2016.

\bibitem{McMeekin2017}
Peter McMeekin, Philip White, Martin~A. James, Christopher~I. Price, Darren Flynn, and Gary~A. Ford.
\newblock Estimating the number of {UK} stroke patients eligible for endovascular thrombectomy.
\newblock {\em European Stroke Journal}, 2:319--326, 12 2017.

\bibitem{SSNAP2023}
Sentinel Stroke National~Audit Programme.
\newblock Ssnap annual report 2023, 2023.

\bibitem{Hausegger2001}
Klaus~A Hausegger, Peter Schedlbauer, Hannes~A Deutschmann, and Kurt Tiesenhausen.
\newblock Complications in endoluminal repair of abdominal aortic aneurysms.
\newblock {\em European Journal of Radiology}, 39:22--33, 2001.

\bibitem{Rudnick1995}
M.~R. Rudnick, S.~Goldfarb, L.~Wexler, P.~A. Ludbrook, M.~J. Murphy, E.~F. Halpern, J.~A. Hill, M.~Winniford, M.~B. Cohen, and D.~B. VanFossen.
\newblock Nephrotoxicity of ionic and nonionic contrast media in 1196 patients: A randomized trial.
\newblock {\em Kidney International}, 47:254--261, 1995.

\bibitem{Klein2009}
Lloyd~W Klein, Donald~L Miller, Stephen Balter, Warren Laskey, David Haines, Alexander Norbash, Matthew~A Mauro, and James~A Goldstein.
\newblock Occupational health hazards in the interventional laboratory: Time for a safer environment.
\newblock {\em Society of Interventional Radiology}, 250:538--544, 2 2009.

\bibitem{Ho2007}
Pei Ho, Stephen~W.K. Cheng, P.~M. Wu, Albert~C.W. Ting, Jensen~T.C. Poon, Clement~K.M. Cheng, Joseph~H.M. Mok, and M.~S. Tsang.
\newblock Ionizing radiation absorption of vascular surgeons during endovascular procedures.
\newblock {\em Journal of Vascular Surgery}, 46:455--459, 9 2007.

\bibitem{Madder2017}
Ryan~D. Madder, Stacie VanOosterhout, Abbey Mulder, Matthew Elmore, Jessica Campbell, Andrew Borgman, Jessica Parker, and David Wohns.
\newblock Impact of robotics and a suspended lead suit on physician radiation exposure during percutaneous coronary intervention.
\newblock {\em Cardiovascular Revascularization Medicine}, 18:190--196, 4 2017.

\bibitem{Crinnion2022}
William Crinnion, Ben Jackson, Avnish Sood, Jeremy Lynch, Christos Bergeles, Hongbin Liu, Kawal Rhode, Vitor~Mendes Pereira, and Thomas~C. Booth.
\newblock Robotics in neurointerventional surgery: a systematic review of the literature.
\newblock {\em Journal of neurointerventional surgery}, 14:539--545, 6 2022.

\bibitem{Riga2010}
Celia~V. Riga, Nicholas~J.W. Cheshire, Mohamad~S. Hamady, and Colin~D. Bicknell.
\newblock The role of robotic endovascular catheters in fenestrated stent grafting.
\newblock {\em Journal of Vascular Surgery}, 51:810--820, 4 2010.

\bibitem{Mofatteh2021}
Mohammad Mofatteh.
\newblock Neurosurgery and artificial intelligence.
\newblock {\em AIMS Neuroscience}, 8:477--495, 2021.

\bibitem{Jackson2023}
Benjamin Jackson, William Crinnion, Mikel De~Iturrate Reyzabal, Harry Robertshaw, Christos Bergeles, Kawal Rhode, and Thomas Booth.
\newblock Comparative verification of control methodology for robotic interventional neuroradiology procedures.
\newblock {\em International Journal of Computer Assisted Radiology and Surgery}, 7 2023.

\bibitem{Sarker2021}
Iqbal~H. Sarker.
\newblock Machine learning: Algorithms, real-world applications and research directions.
\newblock {\em SN Computer Science}, 2, 5 2021.

\bibitem{Fatima2017}
Meherwar Fatima and Maruf Pasha.
\newblock Survey of machine learning algorithms for disease diagnostic.
\newblock {\em Journal of Intelligent Learning Systems and Applications}, 09:1--16, 2017.

\bibitem{Silahtaroglu2021}
Gökhan Silahtaroğlu and Nevin Yılmaztürk.
\newblock Data analysis in health and big data: A machine learning medical diagnosis model based on patients’ complaints.
\newblock {\em Communications in Statistics - Theory and Methods}, 50:1547--1556, 2021.

\bibitem{Robertshaw2023}
Harry Robertshaw, Lennart Karstensen, Benjamin Jackson, Hadi Sadati, Kawal Rhode, Sebastien Ourselin, Alejandro Granados, and Thomas~C. Booth.
\newblock Artificial intelligence in the autonomous navigation of endovascular interventions: a systematic review.
\newblock {\em Frontiers in Human Neuroscience}, 17, 8 2023.

\bibitem{Sutton1998}
Richard~S Sutton and Andrew~G Barto.
\newblock {\em Introduction to Reinforcement Learning}.
\newblock MIT Press, 1st edition, 1998.

\bibitem{Adams2022}
Stephen Adams, Tyler Cody, and Peter~A. Beling.
\newblock A survey of inverse reinforcement learning.
\newblock {\em Artificial Intelligence Review}, 55:4307--4346, 8 2022.

\bibitem{Chi2018}
Wenqiang Chi, Jindong Liu, Mohamed E. M.~K. Abdelaziz, Giulio Dagnino, Celia Riga, Colin Bicknell, and Guang-Zhong Yang.
\newblock Trajectory optimization of robot-assisted endovascularcatheterization with reinforcement learning.
\newblock In {\em 2018 IEEE/RSJ International Conference on Intelligent Robots and Systems (IROS)}, pages 3875--3881. IEEE, 8 2018.

\bibitem{Behr2019}
Tobias Behr, Tim~Philipp Pusch, Marius Siegfarth, Dominik Hüsener, Tobias Mörschel, and Lennart Karstensen.
\newblock Deep reinforcement learning for the navigation of neurovascular catheters.
\newblock {\em Current Directions in Biomedical Engineering}, 5:5--8, 9 2019.

\bibitem{Kweon2021}
Jihoon Kweon, Kyunghwan Kim, Chaehyuk Lee, Hwi Kwon, Jinwoo Park, Kyoseok Song, Young~In Kim, Jeeone Park, Inwook Back, Jae~Hyung Roh, Youngjin Moon, Jaesoon Choi, and Young~Hak Kim.
\newblock Deep reinforcement learning for guidewire navigation in coronary artery phantom.
\newblock {\em IEEE Access}, 9:166409--166422, 2021.

\bibitem{Ng2000}
Andrew~Y Ng and Stuart Russel.
\newblock Algorithms for inverse reinforcement learning.
\newblock pages 663--670. Morgan Kaufmann Publishers Inc., 2000.

\bibitem{Haarnoja2018}
Tuomas Haarnoja, Aurick Zhou, Pieter Abbeel, and Sergey Levine.
\newblock Soft actor-critic: Off-policy maximum entropy deep reinforcement learning with a stochastic actor.
\newblock In Jennifer Dy and Andreas Krause, editors, {\em Proceedings of the 35th International Conference on Machine Learning}, volume~80 of {\em Proceedings of Machine Learning Research}, pages 1861--1870. PMLR, 10--15 Jul 2018.

\bibitem{Karstensen2023}
Lennart Karstensen, Jacqueline Ritter, Johannes Hatzl, Floris Ernst, Jens Langejürgen, Christian Uhl, and Franziska Mathis-Ullrich.
\newblock Recurrent neural networks for generalization towards the vessel geometry in autonomous endovascular guidewire navigation in the aortic arch.
\newblock {\em International Journal of Computer Assisted Radiology and Surgery}, 2023.

\bibitem{Faure2012}
François Faure, Christian Duriez, Hervé Delingette, Jérémie Allard, Benjamin Gilles, Stéphanie Marchesseau, Hugo Talbot, Hadrien Courtecuisse, Guillaume Bousquet, Igor Peterlik, and Stéphane Cotin.
\newblock {\em SOFA, a Multi-Model Framework for Interactive Physical Simulation}.
\newblock Springer Berlin Heidelberg, 2012.

\bibitem{Duriez2006}
C.~Duriez, S.~Cotin, J.~Lenoir, and P.~Neumann.
\newblock New approaches to catheter navigation for interventional radiology simulation.
\newblock {\em Computer Aided Surgery}, 11:300--308, 1 2006.

\bibitem{Blender2018}
Blender~Online Community.
\newblock {\em Blender - a 3D modelling and rendering package}.
\newblock Blender Foundation, Stichting Blender Foundation, Amsterdam, 2018.

\bibitem{Karstensen2022}
Lennart Karstensen, Jacqueline Ritter, Johannes Hatzl, Torben Pätz, Jens Langejürgen, Christian Uhl, and Franziska Mathis-Ullrich.
\newblock Learning-based autonomous vascular guidewire navigation without human demonstration in the venous system of a porcine liver.
\newblock {\em International Journal of Computer Assisted Radiology and Surgery}, 17:2033--2040, 11 2022.

\bibitem{Ziebart2008}
Brian~D Ziebart, Andrew Maas, J~Andrew Bagnell, and Anind~K Dey.
\newblock Maximum entropy inverse reinforcement learning.
\newblock pages 1433--1438. AAAI Press, 2008.

\bibitem{Mankins1995}
John Mankins.
\newblock Technology readiness level - a white paper, 4 1995.

\bibitem{Beaman2023}
Charles Beaman, Ayushi Gautam, Catherine Peterson, Naoki Kaneko, Luciano Ponce, Hamidreza Saber, Kasra Khatibi, Jose Morales, David Kimball, Jacob~Ridge Lipovac, Kazim~H. Narsinh, Amanda Baker, M.~Travis Caton, Eric~R. Smith, May Nour, Viktor Szeder, Reza Jahan, Geoffrey~P. Colby, Branden~J. Cord, Daniel~L. Cooke, Satoshi Tateshima, Gary Duckwiler, and Ben Waldau.
\newblock Robotic diagnostic cerebral angiography: A multicenter experience of 113 patients.
\newblock {\em Journal of NeuroInterventional Surgery}, 2023.

\end{thebibliography}

\end{document}